\documentclass[12pt]{amsart}

\usepackage[margin=1in]{geometry}
\usepackage{amsfonts,amsmath,amssymb}
\usepackage{amsaddr}
\usepackage{graphicx}
\graphicspath{{figures/}}
\usepackage{wrapfig}
\usepackage{subcaption}
\usepackage{epstopdf}
\usepackage{algorithmic}
\usepackage{dsfont}
\usepackage{booktabs}
\usepackage{url}

\def\minimize{\operatorname*{minimize}}

\usepackage{enumitem}
\newcommand{\R}{\mathds{R}}

\newtheorem{remark}{Remark}

\title{Spline Parameterization of Neural Network Controls for Deep Learning}

\author{Stefanie G{\"u}nther, Will Pazner}
\address{Center for Applied Scientific Computing, Lawrence Livermore National Laboratory, Livermore CA}
\author{Dongping Qi}
\address{Center for Applied Mathematics, Cornell University, Ithaca NY}

\usepackage{amsopn}

\begin{document}

\maketitle

\begin{abstract}
Based on the continuous interpretation of deep learning cast as an optimal control problem, this paper investigates the benefits of employing B-spline basis functions to parameterize neural network controls across the layers. Rather than equipping each layer of a discretized ODE-network with a set of trainable weights, we choose a fixed number of B-spline basis functions whose coefficients are the trainable parameters of the neural network. Decoupling the trainable parameters from the layers of the neural network enables us to investigate and adapt the accuracy of the network propagation separated from the optimization learning problem. We numerically show that the spline-based neural network increases robustness of the learning problem towards hyperparameters due to increased stability and accuracy of the network propagation. Further, training on B-spline coefficients rather than layer weights directly enables a reduction in the number of trainable parameters.

\end{abstract}

\section{Introduction}

Advances in algorithms and computational resources have recently fueled many practical applications of deep learning, including classical machine learning areas such as image recognition, classification or segmentation \cite{he2016deep, ronneberger2015u} or natural language processing \cite{goldberg2017neural}, as well as applications in the new and rapidly evolving field of scientific machine learning \cite{rackauckas2020universal, HornungEtAl2020, HeyEtAl2020, spears2018deep, roscher2020explainable}.
While successful applications are reported frequently, advancements on the theoretical side that provide guidance to design and steer new deep learning applications are only at an early stage.
Reaching a desired accuracy for new applications is often a delicate and cumbersome task, requiring domain experts to choose between various network architectures and learning models, a process that often involves hundreds of training runs to scan over various hyperparameters, such as network width and depth, type of network layers, learning rate and adaptation, parameter initialization, etc.
The need for a theoretical foundation that provides guidance for designing new deep learning applications is of particular importance in scientific machine learning which often demand greater accuracy and robustness \cite{raissi2019physics}.

A promising pathway to build upon a theoretical baseline is provided by the recently made interpretation of feed-forward residual networks as discretizations of nonlinear dynamical systems \cite{weinan2017proposal, benning2019deep}.
Instead of considering a neural network as a concatenation of discrete network layers whose parameters are learned during training, the continuous interpretation takes the infinitesimal limit of residual layers to yield a network that is described by a system of nonlinear ordinary differential equations (ODEs), and is driven by a continuous control function:
\begin{align}\label{eq:odenet}
  \frac{\partial x(t)}{\partial t} = f(x(t), \theta(t)), \quad \forall t\in (0,T).
\end{align}
Here, $x(t)$ denotes the propagated state of the input data with $x(0) = x_{data}$ at time $t=0$, and the final time $T$ is associated to the depth of the network. $\theta(t)$ denotes a control function that represents the network parameters and is learned during training.
A standard choice for the right-hand side is $f(x(t), \theta(t))$ =  $\sigma(W(t)x(t)+b(t))$, where $\sigma$ is an activation function applied elementwise ($\tanh$, ReLU, etc), $W(t)$ is a linear transformation matrix representing e.g. a dense parameterized matrix or a convolution operation and $b(t)$ is a bias vector.
We write the control function $\theta(t)=(W(t),b(t))$.
The continuous ODE network models the derivative of the network state in the right hand side of an ODE, casting learning into an optimal control problem that optimizes for the control function $\theta(t)$ to achieve the desired network output:
\begin{align} \label{eq:minloss}
  \minimize_{\theta(t)} \quad  \ell(x(T), y_{data})& + \gamma\int_0^T R(\theta(t)) \, \mathrm{d}t \\
  \text{subject to} \quad
  \frac{\partial x(t)}{\partial t} &= \sigma(W(t)x(t) + b(t)), \quad \forall t\in (0,T), \label{eq:ode}\\
  x(0) &= x_{data}.
\end{align}
where $\ell$ denotes a loss function that measures the mismatch between the predicted network output $x(T)$ and the truth for given input-output data $(x,y)_{data}$, and $R$ denotes a regularization term with parameter $\gamma>0$.

To solve the optimal control learning problem, the temporal domain is discretized into distinct time steps (layers), typically equally spaced with $t_i=ih, i=0,\dots, N$ for a step size $h = T/N$, where the network states and parameters are approximated numerically as $x_i \approx x(t_i), W_i \approx W(t_i), b_i \approx b(t_i)$, using a numerical time integration scheme.
For example, a forward Euler time integration scheme to discretize \eqref{eq:odenet} on a uniformly spaced grid
yields
\begin{align}
\label{eq:discrete_odenet_BE}
  x_{i+1} = x_i + h \sigma(W_ix_i + b_i), \quad \forall i=0,\dots N-1,
\end{align}
with step size parameter $h>0$.
In this case, choosing $h=1$ yields the standard ResNet architecture \cite{he2016deep}.
Each time step $t_i$ corresponds to one layer of a network with depth $N$ and layer spacing $h$.
Each layer is associated with one set of network parameters $\theta_i = (W_i, b_i)$ that have to be learned during training.
Hence, increasing the accuracy of the time stepping scheme for example  by reducing the step size $h$ and increasing $N=\frac{T}{h}$ accordingly results in a linear increase in the number of trainable parameters increasing the dimensionality and complexity of the optimization problem.
On the other hand, larger time step sizes may lead to unstable forward propagation with eigenvalues outside of the stability region of the chosen discretization scheme.
When each layer (at time step $t_i$) is coupled to a set of trainable parameters $\theta_i$, finding an appropriate step size $h$ that ensures stable forward (and backward) propagation is therefore challenging.

The main intent of this paper is to explore the benefits of decoupling the discretization of the ODE network states $x(t)$ from the parameterization of the network control function $\theta(t)$.
We propose a spline-based network that parameterizes the control $\theta(t)$ by a set of fixed basis functions:
\begin{align}
  \theta(t) = \sum_{l=-d}^{L-1} \alpha_l B^d_l(t)
\end{align}
whose coefficients are the trainable parameters of the network.
We choose $B_l^d(t)$ to be a B-spline basis function of degree $d$ as defined in Section \ref{sec:splines}.
Instead of equipping each layer at $t_i$, $i=0,\dots,N$ with a set of parameters $\theta_i$ that are learned during training, the spline-based network is trained for the set of $(L+d)$ coefficients $\alpha_l$ where $L$ is defined as the number of ``time knots'' to construct B-spline bases.
The number of coefficients $(L+d)$ can be significantly smaller that the number of layers (i.e. the number of time steps) and, most importantly, can be chosen independent of the time integration scheme used to discretize \eqref{eq:odenet}.
In fact, a given set of network parameters $\{\alpha_l\}$, for example those from a previous training runs, can be readily evaluated on other network architectures.
The ability to evaluate the network control at any point $t\in[0,T]$ for a fixed set of network parameters allows one to investigate, and potentially increase the accuracy of the discretized network propagation, e.g. by re-discretizing \eqref{eq:odenet} with a integration scheme of higher order, or by choosing smaller, or adaptive time step sizes to discretize $[0,T]$.
While the latter entails an increase in the number of layers $N$, favoring very deep networks for accuracy, it does not increase the number of trainable parameters which instead is independent of $N$ and can be tuned to account for a desired approximation power of the network.
Further, parameterizing the network control with spline basis functions achieves smoothness of the network parameters across layers by design,  rendering additional regularization terms on smoothness of the parameters such as minimizing the derivative of $\theta(t)$ across layers as in \cite{haber2017stable} unnecessary.
Instead, the degree of the chosen basis functions $B^d_l(t)$ controls the regularity of $\theta(t)$.
Avoiding discontinuities in the controls contributes to more accurate and stable forward (and backward) propagation through the discretized network such that small perturbations to the network input yield bounded perturbations of the network output - a requisite for successful training where controls are updated based on the network output sensitivities.
We expect, and numerically verify, that the spline-based network therefore encounters greater regularity in terms of robustness with respect to hyperparameters, allowing for a greater range of hyperparameters that yield successful training.

B-spline basis functions, which have seen success when applied to the isogemetric analysis (IGA) of numerical partial differential equations \cite{Cottrell2009}, provide a natural parameterization for the control function for a number of reasons.
Firstly, the spline basis has local support, which is essential for computational efficiency.
Additionally, the lowest-order (piecewise linear) B-splines can be used to recover a standard residual network if the spline knots are chosen to be coincident with the time step points, and increasing the degree of the B-spline functions increases the regularity of the control function, allowing the smoothness of the parameterization to be considered as a hyperparameter.
Finally, the spline basis is hierarchical with respect to the polynomial degree, and thus is well-suited for adaptive methods and local refinement, a subject of future investigation.

\subsection{Related work}

In \cite{queiruga2020continuous}, a family of \textit{continuous-in-depth} generalizations of residual networks was considered.
These networks, termed ContinuousNets, can be integrated using any standard time integration method, e.g.\ Runge--Kutta methods, and can replace ResNet blocks in problem-specific neural architectures.

In \cite{massaroli2020dissecting}, Massaroli and colleagues considered two discretization approaches for the optimization problem posed in infinite-dimensional functional space.
One method, referred to in that work as a spectral or Galerkin discretization, is based on parameterizing the control function with an \textit{orthogonal basis}, and then truncating the resulting series to obtain a finite-dimensional problem.
An alternative method considered in the same work is the piecewise-constant approximation of the control function, which is referred to as \textit{stacked neural ODEs}.

\section{SpliNet: Weight parameterization using spline basis functions} \label{sec:splines}

In order to decouple the trainable network parameters from the time-domain discretization of the ODE-based network \eqref{eq:odenet}, we introduce a class of networks termed ``SpliNet'' that parameterize the network weights and biases $\theta(t) = (W(t), b(t))$ using B-spline basis functions:
    \begin{align}
        W(t) = \sum_{l=-d}^{L-1} \omega_l B^d_l(t),\\
        b(t) = \sum_{l=-d}^{L-1} \beta_l B^d_l(t),
    \end{align}
    where the coefficients $\omega_l, \beta_l$ are the trainable parameters of the neural network.
    $\omega_l, \beta_l$ have the same dimensions as $W(t), b(t)$ respectively, and $B^d_l(t)$'s are $(L+d)$ fixed one-dimensional B-spline basis functions of degree $d\geq 1$ as defined below.

  \subsection{Definition of the B-spline basis functions}\label{sec:bspline_definition}
\begin{figure}[!htb]
    \centering
    \begin{subfigure}[b]{0.32\textwidth}
        \centering
        \includegraphics[trim=30 0 30 0,clip,width=\textwidth]{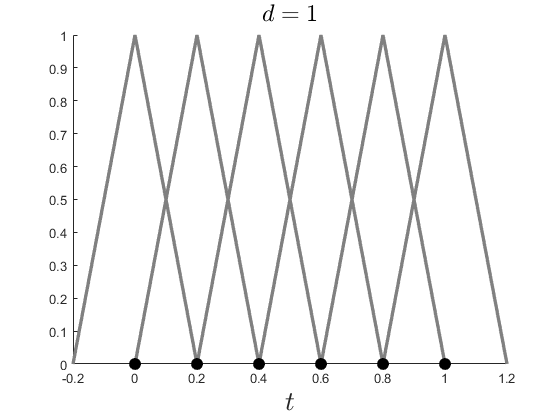}
    \end{subfigure}
    \begin{subfigure}[b]{0.32\textwidth}
        \centering
        \includegraphics[trim=30 0 30 0,clip,width=\textwidth]{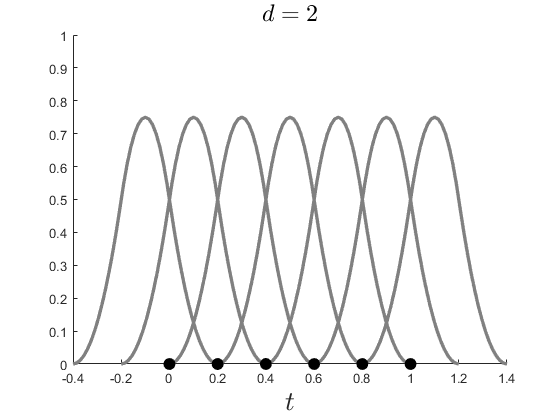}
    \end{subfigure}
    \begin{subfigure}[b]{0.32\textwidth}
        \centering
        \includegraphics[trim=30 0 30 0,clip,width=\textwidth]{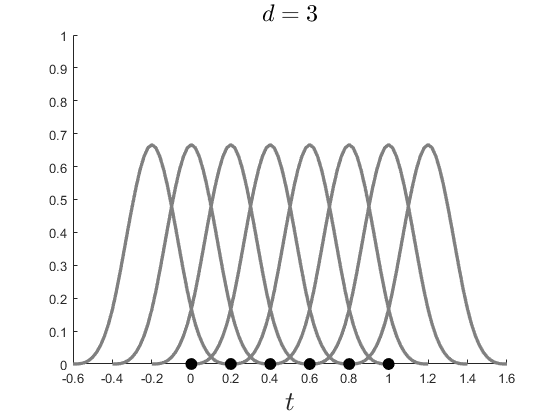}
    \end{subfigure}
    \caption{B-Spline basis functions for degree $d = 1,2,3$, defined on $L+1=6$ uniformly spaced knots $\tau_l \in [0,1]$. As the degree increases, more basis functions are needed to interpolate a function on [0,1] and each basis becomes smoother.}
    \label{fig:spline_basis}
\end{figure}

  To define the one-dimensional B-spline basis functions, consider an equally spaced time grid of $(L+1)$ knots $0=\tau_0<\tau_1<\cdots<\tau_L=T$.
  Each B-spline basis function $B^d_l(t)$ is defined as a polynomial of degree $d$ with local support in $[\tau_l,\tau_{l+d+1}]$ whose first $(d-1)$ derivatives are continuous across the knots.
  The basis functions can be constructed recursively (cf.\ the Cox–de Boor recursion formula \cite{kincaid2009numerical}), starting with a degree-zero B-spline $B^0_l(t)$, defined as the indicator function on $[\tau_l,\tau_{l+1})$
    \begin{align}
         B^0_l(t) = \begin{cases}
           1, & \text{if} \quad t\in[\tau_l,\tau_{l+1}), \\
           0, & \text{otherwise}.
        \end{cases}
    \end{align}
    The higher-degree B-splines are defined by the recursion
    \begin{align} \label{eq:recursive_bspline}
           B^d_l(t) = \left(\frac{t-\tau_l}{\tau_{l+d}-\tau_l}\right) B^{d-1}_l(t) +
                   \left(\frac{\tau_{l+d+1}-t}{\tau_{l+d+1}-\tau_{l+1}}\right) B^{d-1}_{l+1}(t)
     \end{align}
    for $d \geq 1$.
    Figure \ref{fig:spline_basis} shows examples for B-spline basis functions of degree $d=1,2,3$ defined on $[0,1]$.
The resulting entries of the weight and bias functions $W(t), b(t)$ are piecewise continuous polynomials of degree $d$ whose derivatives at knots are continuous up to the $(d-1)$-th derivative.

\subsection{Network propagation using spline-based parameters}

Choosing a finite number of basis functions to parameterize the network control reduces the infinite dimensional learning problem of finding functions $W(t), b(t)$ as in \eqref{eq:minloss} to the finite dimensional problem of learning $(L+d)$ coefficients $\omega_l, \beta_l$ by solving
    \begin{align}
    \minimize_{\omega_l, \beta_l, l=-d,\dots,L-1} \quad  \ell(x(T), y_{data})& + \gamma\int_0^T R(\theta(t)) \, \mathrm{d}t \\
  \text{subject to} \quad
  \frac{\partial x(t)}{\partial t} &= \sigma\left(\sum_{l=-d}^{L-1} \left(\omega_l x(t) + \beta_l \right)B^d_l(t)\right), \quad \forall t\in (0,T),\\  x(0) &= x_{data}.
\end{align}

\begin{figure}
  \centering
  \includegraphics[trim=30 10 30 35, clip,width=0.6\textwidth]{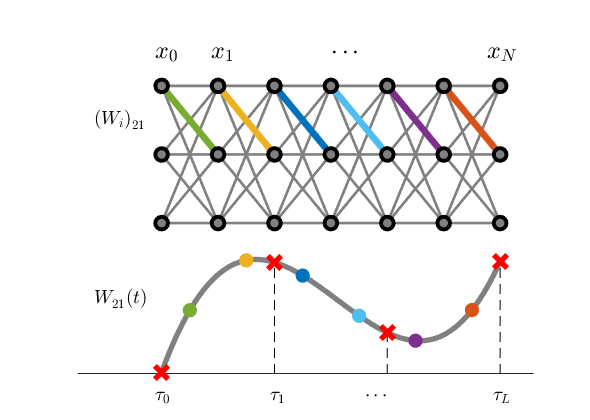}
  \caption{An illustration of SpliNet with $N=6$ and $L+1 = 4$. $W$ is always a $3\times 3$ matrix and one of its entries $W_{21}^{ }$ is highlighted using colored edges. These entries can be evaluated at the corresponding colored dots using the B-Spline function shown underneath. The red crosses are knots of the spline function, corresponding to $\tau_l$'s. SpliNet applies this parameterization to each entry of $W$ and $b$.}
  \label{fig:splinet}
\end{figure}

To train and evaluate a SpliNet, a temporal integration scheme for the network ODE needs to be chosen, obtaining, for example, \eqref{eq:discrete_odenet_BE} in case of a forward Euler time integration scheme.
Each discrete time step $t_i \in [0,T], i=1,\dots,N$ then corresponds to one layer of the discretized neural network where weights and biases $W(t_i), b(t_i)$ need to be computed by summing over the coefficients multiplied by spline basis functions evaluated at $t_i$.
Since each basis function $B^d_l(t)$ has local support in $[\tau_l, \tau_{l+d+1})$, only $(d+1)$ basis functions are non-zero at any time step $t_i\in [0,T]$. In particular, given a time-point $t_i$ which lies in the interval $[\tau_{k_0}, \tau_{k_0+1})$, for one specific $k_0 \geq 0$, only $B^d_{k_0-d}(t_i), \dots, B^d_{k_0}(t_i)$ are non-zero and contribute to the evaluation of the weights and biases at $t_i$:
    \begin{align}
    \label{eq:wb_bspline_compact}
        W(t_i) &= \sum_{l=k_0-d}^{k_0} \omega_l B^d_l(t_i), \quad \text{for} \quad t_i \in [\tau_{k_0},\tau_{k_0+1}), \\
        b(t_i) &= \sum_{l=k_0-d}^{k_0} \beta_l\, B^d_l(t_i), \quad \text{for} \quad t_i \in [\tau_{k_0},\tau_{k_0+1}),
    \end{align}
for the current network control parameters $\omega_l, \beta_l$.
After gathering the weight matrix and bias at $t_i$, the corresponding $i$-th layer can be readily applied to the network states $x_i$.
Figure \ref{fig:splinet} depicts an example of a discretized neural network of dense layers of width $3$, where each entry of the weight matrix at each layer $t_i$ results from evaluating a spline function that represents the weights over time.

A spline-based neural network can be integrated with modern deep learning software modules such as PyTorch \cite{NEURIPS2019_9015} or Tensorflow \cite{tensorflow2015-whitepaper} with the only additional step being the summation for weight and bias elements before applying the layer. Software libraries for B-spline basis functions are readily available, e.g. through the python-based SciPy package \cite{2020SciPy-NMeth}, among others.
Similar to standard ODE networks that parameterize each time step with a set of network parameters, a SpliNet can serve as a block within a larger network composed of sub-networks of different architectures.

\begin{remark}
    When B-spline knots $\tau_l$ align with time-discretization points $t_i$ (layers), we note that the first-order basis functions yield $B^1_l(t_i) = \delta_{li}$ thus $W(t_i) = \omega_i$.
    Therefore, a degree-one SpliNet with $L+1 = N$ is equivalent to a standard ODE network. Choosing further $h=1$ yields a ResNet architecture.
\end{remark}

\begin{figure}[!htb]
    \centering
    \begin{subfigure}[b]{0.32\textwidth}
        \centering
        \includegraphics[trim=80 0 70 0,clip,width=\textwidth]{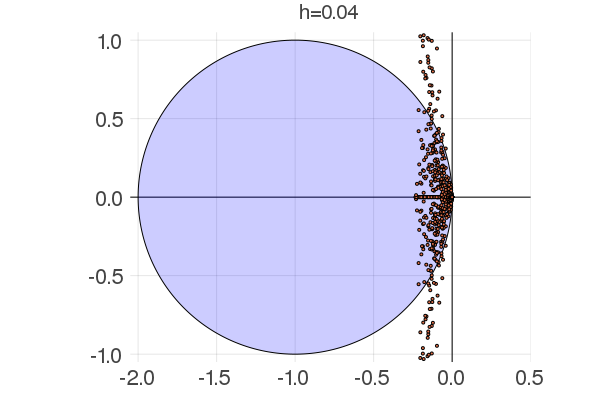}
    \end{subfigure}
    \begin{subfigure}[b]{0.32\textwidth}
        \centering
        \includegraphics[trim=80 0 70 0,clip,width=\textwidth]{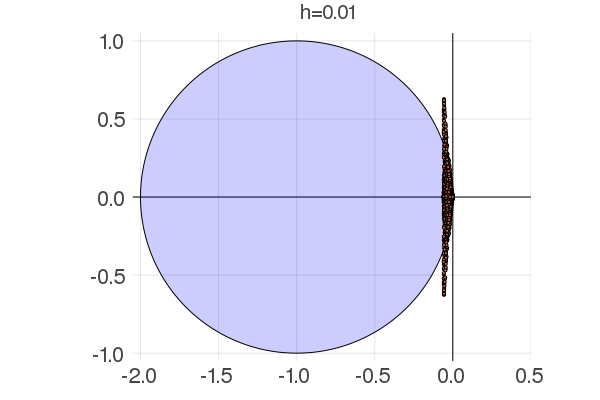}
    \end{subfigure}
    \begin{subfigure}[b]{0.32\textwidth}
        \centering
        \includegraphics[trim=80 0 70 0,clip,width=\textwidth]{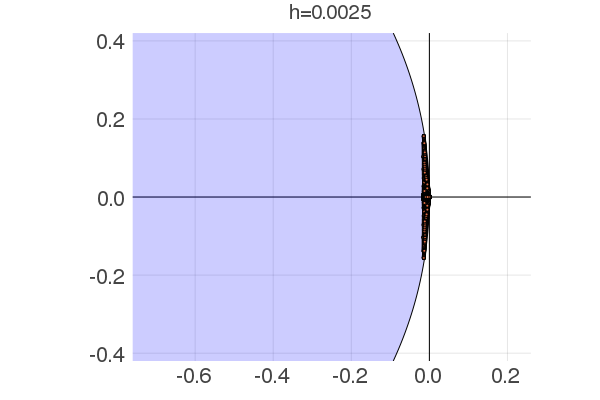}
    \end{subfigure}
   \caption{Scatter plot of the spectra in \eqref{eq:stable_region} of a trained network with different step sizes $h$.
    The shaded circle is the stability region of the forward Euler integration scheme, i.e. $\{z:|1+z|\leq 1\}$.
    Although the network structure ensures that $W_i$'s eigenvalues only have negative real parts, the step size $h$ needs to be adapted to scale them into the stability region.}

   \label{fig:stability}
   \end{figure}

The SpliNet allows one to investigate and adapt accuracy and stability of the network propagation for a fixed set of network parameters.
Accuracy can be increased by adapting the time step size $h$ and increasing the number of time steps $N$.
Concerning stability, it is natural to require that the network propagation steps lie in the stability region of the chosen time integration scheme.
For a forward Euler discretization, the stability condition is given by
   \begin{equation}
   \label{eq:stable_region}
       \Lambda\left(h \,\text{diag}\left[\sigma'\left(W(t_i)x(t_i)+b(t_i)\right)\right]W(t_i)\right) \subset \{z\in\mathbb{C}:|1+z|\leq 1\}, \quad \forall \, i=0,\dots,N
   \end{equation}
  where $\Lambda$ denotes the spectrum of the operator.
   In \cite{haber2017stable}, Ruthotto et al. proposed several variations of ODE-based neural networks which enforce \eqref{eq:stable_region} on the continuous level through constraints on $W(t)$.
   For example, replacing $W(t)$ by $[W(t) - W(t)^\intercal]$ yields anti-symmetric weights such that eigenvalues of the Jacobians are purely imaginary thus the continuous dynamics are stable.
   Furthermore, a negative multiple of the identity $-\gamma I$ with $\gamma > 0$ can be added to the weights to assure that the eigenvalues have negative real parts.
   For the discretized problem, a SpliNet allows to adapt the step size $h$ to scale the eigenvalues into the stability region $\{z\in\mathbb{C}: |1+z| \leq 1\}$, for fixed network controls.
   As an example, in Figure \ref{fig:stability} we parameterize an anti-symmetric weights network using SpiNet and show the distribution of eigenvalues at all the layers for a trained network using the $\sin(x)$ test case (see Section \ref{sec:sine_example}).
   Furthermore, in Section \ref{sec:numerical}, a numerical experiment (see Figure \ref{fig:sine_dttest}) is conducted to justify the stability and accuracy of SpliNet as the step size $h$ tends to zero.

   \begin{remark}
    We note that while \cite{haber2017stable} has discussed the benefits of stable forward propagation, numerical results suggest that for some problems, successful training with ODE networks can be obtained even for operators whose spectrum is not entirely contained in the stability region of the time integration scheme.
    In the test cases considered in Section \ref{sec:numerical}, we do not enforce this property using the antisymmetric weight matrices, and instead allow for general weight matrices.
   \end{remark}

\subsection{Backpropagation using spline-based parameters}

    To solve the training optimization problem, backpropagation is typically employed to compute the gradient of the loss at $x_N^{ }$ with respect to the trainable parameters $\omega_l, \beta_l$ and perform a gradient step to update the parameters.
    Backpropagation is equivalent to the adjoint method commonly used in ODE/PDE-constrained optimization where an adjoint differential equation is solved backwards in time propagating derivatives through the time domain.
    In the discretized setting, the adjoint method accumulates derivatives backwards from the network output and loss evaluation to the network input using the chain rule.
    Here, we briefly discuss how the adjoint method (backpropagation) is used to compute the gradient with respect to the B-spline coefficients, i.e.\ the real trainable parameters in a SpliNet.
    We derive the gradient for network inputs that are vectors, however, a similar derivation can be obtained for a tensor input.

    Let $\Phi(x_i; W_i, b_i)$ denote a general layer-to-layer transformation, i.e. the right hand side of the time-discretized ODE, such as for example $\Phi(x_i;W_i,b_i) = x_i + h\sigma(W_i x_i + b_i)$ for a forward Euler discretization. The adjoint method equips each state $x_i$ with an adjoint state $z_i$ that is given by partial derivatives with respect to the network states:
    \begin{align}
        z_N^{ } &= \frac{\partial \ell}{\partial x_N^{ }}(x_N^{ },y_{data})^\intercal \label{eq:last_adjoint}, \\
        z_i &= \frac{\partial \Phi}{\partial x_i}(x_i;W_i,b_i)^\intercal z_{i+1}, \quad \forall\, i = N-1,\cdots,0. \label{eq:adjoint}
    \end{align}
    Using the adjoint variables, the gradient of loss function with respect to B-Spline coefficients can be computed exploiting linearity between $W$ and $\omega_l$ in \eqref{eq:wb_bspline_compact}:
    \begin{align}
        \frac{\partial \ell}{\partial \omega_l} &= \sum_{\substack{t_i\, \in \\ [\tau_l,\tau_{l+d+1})}}\frac{\partial \Phi}{\partial \omega_l}(x_i;W_i,b_i)^\intercal z_{i+1}, \label{eq:gradients_1} \\
        \frac{\partial \ell}{\partial \beta_l} &= \sum_{\substack{ t_i\, \in \\ [\tau_l, \tau_{l+d+1})}}\frac{\partial \Phi}{\partial \beta_l}(x_i;W_i,b_i)^\intercal z_{i+1}. \label{eq:gradients_2}
    \end{align}
    In contrast to standard control parameterization where derivatives with respect to $W_i$ are desired, an additional step of computing the gradients with respect to the spline coefficients are necessary, i.e. summing gradients over all time steps $t_i$ that lie in $[\tau_l, \tau_{l+d+1}]$.
   When the linear transformation $W_i$ is a matrix (and the state variable $x_i$ is a vector), the right-hand sides of \eqref{eq:gradients_1} and \eqref{eq:gradients_2} are given by
    \begin{align}
    \label{eq:coeff_gradients}
        \frac{\partial \Phi}{\partial \omega_l}(x_i;W_i,b_i)^\intercal z_{i+1} &= h B_l^d(t_i)[\sigma'(W_i\, x_i + b_i) \odot z_{i+1}]x_i^\intercal, \\
        \frac{\partial \Phi}{\partial \beta_l}(x_i;W_i,b_i)^\intercal z_{i+1}  &= h B_l^d(t_i)[\sigma'(W_i\, x_i + b_i) \odot z_{i+1}],
    \end{align}
    where $\odot$ is the Hadamard product.

    Instead of first solving the adjoint equations \eqref{eq:adjoint} backwards for $i=N-1, \dots, 0$ and then summing over time steps as in \eqref{eq:gradients_1} and \eqref{eq:gradients_2} to compute the gradient, the contribution to the gradient from each $t_i$ can also be accumulated during backpropagation of the adjoint variables, updating the gradient at each $i=N-1, \dots, 0$ with
    \begin{align}
    \label{eq:gradient_alternative}
        \frac{\partial \ell}{\partial \omega_l} \, \longleftarrow \, \frac{\partial \ell}{\partial \omega_l} + \frac{\partial \Phi}{\partial \omega_l}(x_i;W_i,b_i)^\intercal z_{i+1}, \quad   \forall \, l = k_0-d,
        \cdots,k_0,
    \end{align}
    for $k_0$ such that $t_i \in [\tau_{k_0},\tau_{k_0+1})$, and similarly for $\frac{\partial \ell}{\partial \beta_l}$.

    Modern deep learning libraries such as PyTorch and Tensorflow utilizes automatic differentiation (AD) to realize backward propagation. The user is only required to implement a forward step to apply the given layer, while the corresponding backward step to update the adjoint variables and accumulate the gradient is performed automatically in the background by moving backwards over the computational graph.
    Compared to a standard ResNet or neural ODE network, the forward propagation of a SpliNet involves the additional step of constructing $W_i$ locally through \eqref{eq:wb_bspline_compact} summing over $d$ network parameters and B-splines at each time step during forward propagation. Hence, $d$ Jacobian-vector products (instead of one) will be performed during backpropagation according to \eqref{eq:gradient_alternative}, either manually implemented or automatically obtained through AD.

    A continuous ODE network favors networks with a large number of time steps (layers) in order to ensure stable propagation and resolve the ODE network dynamics.
    Considering backpropagation, however, the network state at each layer needs to be stored during forward propagation in order to evaluate the partial derivatives of $\Phi$, which induces greater memory requirements.
    To resolve this, checkpointing techniques can be employed which trade memory requirement against computational costs, as for example utilized in ANODE \cite{gholami2019anode}. In this approach, only certain network states are stored during forward propagation while others are recomputed when needed, starting from those ``checkpoints''. An optimal distribution of network states is set in the \textit{revolve} algorithm, as is commonly used in PDE-constrained optimization \cite{griewank2000algorithm}.

\section{Learning the time scale}

\def\tref{\hat{t}}
\def\xref{\hat{x}}
\def\thref{\hat{\theta}}

The system of ordinary differential equations \eqref{eq:odenet} is posed for a pseudo-time variable $t \in (0,T)$, where $t$ represents a continuous analogue of the network depth in a discrete residual network.
Note that a simple change of variables $\tref \mapsto \lambda t$ can recast \eqref{eq:odenet} to be defined over the \textit{reference domain}, $\tref \in (0,1)$,
\begin{align} \label{eq:odenet_refdomain}
    \frac{\partial \xref(\tref)}{\partial\tref} = \lambda f(\xref(\tref), \thref(\tref)),
\end{align}
where $\lambda$ is a \textit{scaling parameter} that determines the time scale of the problem, $\xref(\tref) = x(t)$, and $\thref(\tref) = \theta(t)$.
In other works, this scaling parameter is considered to be a tuning parameter to be determined during hyperparameter optimization \cite{queiruga2020continuous}.
Presently, we also consider incorporating this parameter into the learning process.

Consider the standard choice of right-hand side,
\[
    f(x(t), \theta(t)) = \sigma(W(t)x(t) + b(t)),
\]
for linear transformation $W(t)$, bias vector $b(t)$, and activation function $\sigma$.
The control function $\theta(t)$ is typically the pair $(W(t),b(t))$.
Note that if $\sigma$ is homogeneous with degree 1 (i.e.\ $\sigma(\alpha v) = \alpha \sigma(v)$, satisfied for example by the identity and ReLU functions), then
\[
    \sigma(\hat{W}(\tref)\xref(\tref) + \hat{b}(\tref))
    = \lambda \sigma(W(t)x(t) + b(t)),
\]
for $\hat{W}(\tref) = \lambda W(t)$, and $\hat{b}(\tref) = \lambda b(t)$.
As a consequence, for \textit{homogeneous} activation functions, the time scale of the problem is determined by the magnitude of the weight and bias coefficients, and is therefore automatically optimized for during the learning process.
In this case, tuning the scaling parameter $\lambda$ has an effect only on the initialization of the weights and biases.

For activation functions that are not homogeneous (including the commonly used sigmoid and $\tanh$ functions), the effect of time scaling is not completely determined by a linear scaling of the weight and bias coefficients.
In this case, instead of considering the choice of time scale as a hyperparameter that must be tuned for each problem, we incorporate the scaling parameter $\lambda$ into the control function $\thref(\tref) = (\lambda, \hat{W}(\tref), \hat{b}(\tref))$, which can be optimized for during the optimization procedure.
This allows for a systematic treatment of the pseudo-time variable, avoiding the ad hoc problem dependent choice of final time $T$.
Numerical results concerning the learning of the time scale are presented in Section \ref{sec:impact-time-scale}.

\section{Numerical results}
\label{sec:numerical}
We demonstrate the benefits of a spline-based network on test cases with increasing complexity.
For all test cases, we discretize the network ODE \eqref{eq:odenet_refdomain} in the reference domain $t\in[0,1]$ on $N$ equally distributed time steps $t_i = ih$, $i=0,\dots,N$ with time step size $h$ using a forward Euler time integration scheme, which allows to directly compare results with a standard ResNet architecture.

\begin{figure}
    \centering
    \includegraphics[width=.6\textwidth]{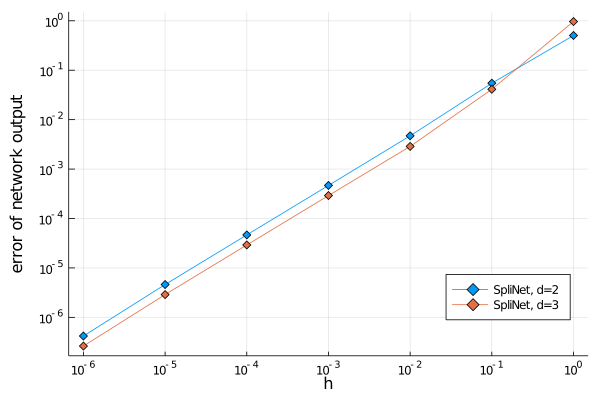}
    \caption{Error of the network output over time step size $h$. Parameterizing network weight with fixed basis functions allows evaluating the same network with any desired accuracy.}
    \label{fig:sine_dttest}
\end{figure}

Before training, we determine the number layers $N$ (time steps) by analyzing the error of the network output with respect to varying time step sizes $h = 1/N$ on randomized network control functions. Figure \ref{fig:sine_dttest} plots the error of the discretized network dynamics compared to the continuous ODE formulation evaluated at final time $T=1$ for various time step sizes $h$, for the $\sin(5x)$ test case as described in Section \ref{sec:sine_example}. We observe first-order convergence of the time integration scheme as expected. Notice that the ability to perform this test is a consequence of parameterizing the network control functions with a finite number of basis function coefficients, allowing for the evaluation of the continuous network at any point in time, independent of the number of layers $N$.

For all test cases, we choose the ADAM optimizer on mini-batches of various sizes. Further, we add a Tikhonov regularization term to the loss function which minimizes the square $L^2$-norm of the network parameters $\omega_l, \beta_l$.
All numerical results are performed using our implementation for B-spline-based neural networks (\cite{splinemlGithub}) written in Julia while utilizing the Julia open-source machine learning library Flux.jl to employ network layers and optimizers, as well as Zygote.jl to incorporate backpropagation through automatic differentiation \cite{innes2018don}.
The implementation of SpliNet provides users with two options, either dense layers (i.e. each $W_i$ being a dense matrix) or convolutional layers (i.e. each $W_i$ being a convolution).
When using convolutional layers, our current SpliNet implementation can handle tensor inputs with rank up to three.

\subsection{Test case descriptions} \label{sec:testcases}
\subsubsection{$\sin(f x)$}\label{sec:sine_example}
This test case aims to learn the sine function $\sin(f x)$ for $x\in [-\pi, \pi]$ with frequencies $f\in \{1,2,5\}$. We utilize $20f$ uniformly distributed data points $x_{data} \in [-\pi,\pi]$ and $y_{data} = \sin(f x_{data})$ for training. We choose a network of dense layers of width $m=4$. At $t=0$, we map the input data $x_{data}$ onto the network width through replication, i.e. $x(0) = [1,1,1,1]^\intercal x_{data}$. The loss function evaluated at $x(T=1)$ computes the $\ell^2$-norm of the averaged network output:
\begin{align}
    l(x(T), y_{data}) = \frac{1}{2} \left(\frac{1}{m}\sum_{j=1}^m x_j(T) - y_{data} \right)^2.
\end{align}
We choose the fixed step size of $h=0.01$ which accounts to $N=100$ layers (time steps) for the spline-based network (compare Figure \ref{fig:sine_dttest}). For the nonlinear activation function $\sigma$, we choose the $\tanh$ function.

\subsubsection{Peaks}
The peaks problem suggested by \cite{haber2017stable} intends to classify 2D grid points into five regions bounded by level sets of a smooth nonlinear function on $[-3,3]^2$.
1000 points $x_{data} \in [-3,3]^2$ are chosen randomly as data.
The goal is to train a network that predicts the correct level sets for new points, hence labels $y_{data}$ are unit vectors in $\R^5$ that indicate the corresponding level set for $x_{data}$.
The nonlinear activation is ReLu function $\max\{0,x\}$.
The weights and biases $W$, $b$ are dense matrices and vectors respectively, whose elements are parameterized by B-splines.
The first layer maps $x_{data}$ to the network width of $m=5$ through replication, and the last layer performs a softmax function to yield a class probability vector for the network output $x(T=1)$.
The loss function at $T=1$ then compares the predicted class probabilities with the true classes represented by $y_{data}$ using the cross entropy loss.

\subsubsection{Image segmentation test case (Indian Pines)}
The Indian Pines test cases is a soil segmentation problem that aims to classify each pixel of 220 hyperspectral images of a single landscape in Indiana US, into one of 16 types of land-cover (such as alfalfa, corn, soybean, wheat, etc.) \cite{PURR1947}.
For training, we use the spectral bands of $4000$ randomly chosen pixel points $x_{data} \in R^{220}$ together with their corresponding class probability $y_{data}$ being unit vectors in $\R^{16}$. The network architecture consists of convolutional layers with a kernel of width $3$ applied to each spectral band, a dense bias vector, and a ReLU activation ($\sigma(x) = \max\{0,x\}$). The network width matches the input dimension for each pixel, being $m=220$ channels corresponding to the $220$ spectral bands. We choose the identity to map $x_{data}$ onto the network width in the first layer and apply a cross entropy loss function to the softmax at the last layer at $T=1$.

\subsection{Implicit regularization from spline-based network controls}
In order to investigate the regularizing effect of the spline-based network, we perform a hyperparameter search over learning rates $\eta\in[10^{-3},10^{-1}]$, regularization parameters $\gamma \in[10^{-10},10^{-4}]$, amplitudes of randomized initial network parameters in $[10^{-3}, 1]$, and the number of spline-coefficient matrices $(L+d)$ with $L \in \{2,3,\dots,15\}$. For the Indian Pines test case, we increase $L \in\{2,5,10,15,20,25\}$ to account for to complexity of this test case.

We show gathered statistics over $100$ training runs, randomly sampled from the above parameters. In order to investigate the benefits resulting from a spline-based parameterization of the network weights and biases, we compare the SpliNet of degrees $d\in \{1,2,3\}$ with (a) an ODE network as in \eqref{eq:discrete_odenet_BE} (``ODEnet'') where each layer/time step owns one set of trainable weights and biases such that $L=N$ as the corresponding hyperparameter (and $h = T/N$), as well as (b) a standard ResNet where again each layer owns one set of weights and biases and $L=N$ being a hyerparameter, but in contrast to an ODEnet, the ResNet uses a fixed step size of $h=1$.

\begin{figure}
    \centering
    \includegraphics[width=0.49\textwidth]{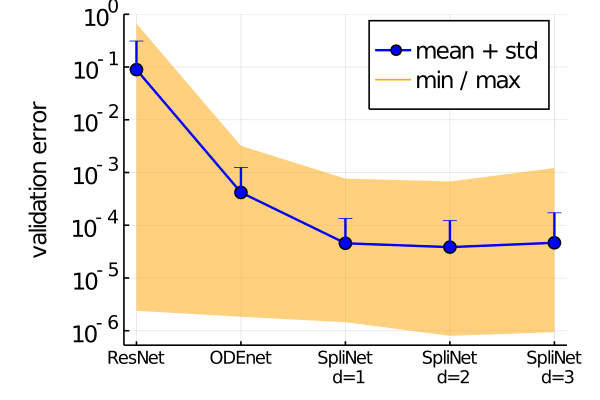}
    \includegraphics[width=0.49\textwidth]{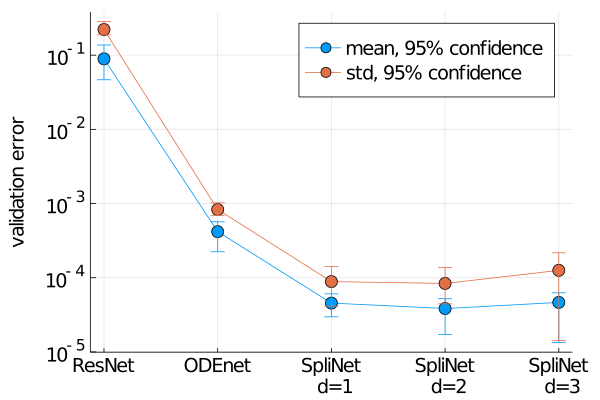}
    \caption{$\sin(2x)$ test case: Statistics of validation error over 100 training runs. Left: Mean with standard deviation error bar, and minimum / maximum band. Right: Standard deviation with 95\% confidence interval.}
    \label{fig:sine_bootstrap_mean_min_max}
\end{figure}

Table \ref{tab:sine_results} displays the gathered statistics for the $\sin(fx)$ test case. First we notice that all networks are able to approximate $\sin(fx)$ with similar accuracy (see column ``min'' for each $f\in \{1,2,5\})$. However, each spline-based network outperforms both the ResNet as well as the ODEnet in terms of mean validation accuracy and standard deviation.
The spline-based networks reduce the mean validation error by two to three orders of magnitude when compared to a ResNet for all $f$, and one order of magnitude over the ODEnet for $f\in\{2,5\}$. Further, the SpliNet is able to drastically reduce the standard deviation as well as maximum error (see column ``std'' and ``max''), which shows the regularization effect of spline-based network.
We visualize the data for $f=2$ in Figure \ref{fig:sine_bootstrap_mean_min_max}, showing mean and standard deviation as well as minimum and maximum validation error (left). To verify that the improvements are statistically significant, we compute $95\%$ confidence intervals for the mean accuracy as well as the standard deviation. Confidence intervals are plotted as error bars in Figure \ref{fig:sine_bootstrap_mean_min_max} (right), which show a significant drop of mean error and standard deviation for the ODE-based network over the ResNet architecture and for all spline-based networks over the ODEnet.
We further observe that the degree of the chosen B-spline basis functions does not yield additional improvement, indicating that degree one basis functions which correspond to piecewise linear weights and bias functions across layers are sufficient to improve robustness towards hyperparameters.

\begin{table}[!htb]
  \center
  \caption{$\sin(fx)$ test case: Validation error statistics over $100$ training runs. The spline-based network consistently reduces the mean validation error as well as standard deviation and maximum validation error.}
  \setlength{\tabcolsep}{12pt}
  \begin{tabular}{@ { } lrrrr @ { }}
    \toprule
		$\sin(x)$ & mean & std & min & max \\
	 \midrule
	 ResNet          & $1.8\times 10^{-2}$ & $4.7\times 10^{-2}$ &  $4.2\times 10^{-7}$ & $2.0\times 10^{-1}$ \\
	 ODEnet          & $3.5\times 10^{-5}$ & $1.2\times 10^{-4}$ &  $2.1\times 10^{-6}$ &  $1.1\times 10^{-3}$ \\
	 SpliNet, $d=1$  & $2.4\times 10^{-5}$ & $2.2\times 10^{-5}$ &  $2.9\times 10^{-6}$ &  $1.3\times 10^{-4}$ \\
	 SpliNet, $d=2$  & $2.2\times 10^{-5}$ & $2.0\times 10^{-5}$ &  $6.8\times 10^{-7}$ &  $1.1\times 10^{-4}$ \\
	 SpliNet, $d=3$  & $2.8\times 10^{-5}$ & $3.4\times 10^{-5}$ &  $2.4\times 10^{-6}$ &  $2.6\times 10^{-4}$ \\
  \midrule
		$\sin(2x)$ & mean & std & min & max \\
	 \midrule
	 ResNet          & $8.9\times 10^{-2}$ & $2.2\times 10^{-1}$ &  $2.3\times 10^{-6}$ & $6.6\times 10^{-2}$ \\
	 ODEnet          & $4.1\times 10^{-4}$ & $8.3\times 10^{-4}$ &  $1.8\times 10^{-6}$ &  $3.2\times 10^{-3}$ \\
	 SpliNet, $d=1$   & $4.5\times 10^{-5}$ & $8.8\times 10^{-5}$ &  $1.4\times 10^{-6}$ & $7.6\times 10^{-4}$ \\
	 SpliNet, $d=2$   & $3.8\times 10^{-5}$ & $8.3\times 10^{-5}$ &  $7.9\times 10^{-7}$ & $6.7\times 10^{-4}$ \\
	 SpliNet, $d=3$   & $4.6\times 10^{-5}$ & $2.2\times 10^{-4}$ &  $9.3\times 10^{-7}$ & $1.2\times 10^{-3}$ \\
	 \midrule
		$\sin(5x)$ & mean & std & min & max \\
	 \midrule
	 ResNet          & $4.7\times 10^{-2}$ & $1.0\times 10^{-1}$ &  $9.5\times 10^{-6}$ & $3.5\times 10^{-1}$ \\
	 ODEnet          & $1.0\times 10^{-2}$ & $2.9\times 10^{-2}$ &  $8.5\times 10^{-6}$ &  $1.8\times 10^{-1}$ \\
	 SpliNet, $d=1$   & $1.2\times 10^{-3}$ & $1.8\times 10^{-3}$ & $6.5\times 10^{-7}$ & $1.8\times 10^{-2}$ \\
	 SpliNet, $d=2$   & $7.4\times 10^{-3}$ & $4.7\times 10^{-2}$ &  $7.2\times 10^{-7}$ & $4.2\times 10^{-1}$ \\
	 SpliNet, $d=3$   & $7.1\times 10^{-3}$ & $4.6\times 10^{-2}$ &  $6.4\times 10^{-7}$ & $4.3\times 10^{-1}$ \\
   \bottomrule
  \end{tabular}
  \label{tab:sine_results}
\end{table}

\begin{figure}[!htb]
    \centering
    \includegraphics[width=0.49\textwidth]{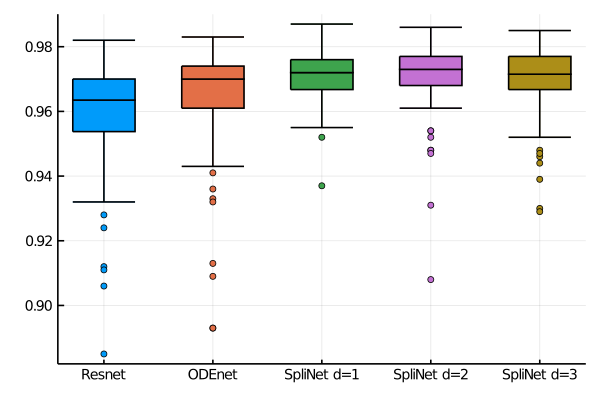}
    \includegraphics[width=0.49\textwidth]{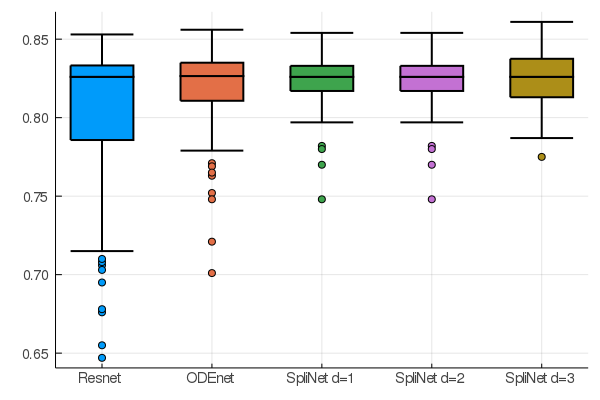}
    \caption{Median accuracy with lower and upper quartiles gathered over 100 training runs for the Peaks test case (left) and the Indian Pines segmentation problem (right).}
    \label{fig:boxplot_peaks_indianpines}
\end{figure}

Similar observation can be obtained from statistics for the Peaks test case as well as the more complex Indian Pines segmentation problem (see Tables \ref{tab:peaks_results}, \ref{tab:indianpines}, and Figure \ref{fig:boxplot_peaks_indianpines}). Most notably, the standard deviation is reduced significantly for the spline-based neural networks, indicating the regularizing effect of a SpliNet as well as tighter bounds of minimum and maximum validation.
\begin{table}[!htb]
  \center
  \caption{Peaks test case. Statistics of validation accuracy gathered over 100 training samples. All spline-based network outperform the ODEnet as well as ResNet architectures in terms of greater mean validation accuracy and tighter standard deviation.}
  \label{tab:peaks_results}
  \begin{tabular}{@ { } lrrrr @ { }}
    \toprule
		   \textit{Peaks}          & mean & std & min & max \\
	 \midrule
	 ResNet                & 95.9\% & 1.7\% & 88.5\% & 98.2\% \\
	 ODEnet                & 96.4\% & 1.6\% & 89.3\% & 98.3\% \\
	 SpliNet, $d=1$        & 97.0\% & 0.8\% & 93.7\% & 98.7\% \\
	 SpliNet, $d=2$        & 97.1\% & 1.1\% & 90.8\% & 98.6\% \\
	 SpliNet, $d=3$        & 97.0\% & 1.0\% & 92.9\% & 98.5\% \\
   \bottomrule
  \end{tabular}
\end{table}

\begin{table}[!htb]
    \centering
    \caption{Indian Pines segmentation problem. Statistics of validation accuracy gathered using 100 training samples.}
    \label{tab:indianpines}
    \begin{tabular}{@ { } lrrrr @ { }}
        \toprule
    	\textit{Indian Pines}  & mean & std & min & max \\
    	 \midrule
    	 ResNet          & 80.1\% & 5.0\% & 64.7\% & 85.3\% \\
    	 ODEnet          & 81.9\% & 2.6\% & 70.1\% & 85.6\% \\
    	 SpliNet, $d=1$  & 82.4\% & 1.8\% & 75.4\% & 85.3\% \\
    	 SpliNet, $d=2$  & 82.3\% & 1.6\% & 74.8\% & 85.4\% \\
    	 SpliNet, $d=3$  & 82.4\% & 1.7\% & 77.5\% & 86.1\% \\
       \bottomrule
    \end{tabular}
\end{table}

\subsection{Reduction of network parameters}
A spline-based neural network decouples the trainable network parameters from the network layers, hence sharing parameters over various layers.
We anticipate that this decoupling allows to reduce the number of parameters without hampering the network's approximation power. Here, we validate this hypothesis on two test cases by investigating mean validation accuracy for training using various numbers $L$ of trainable sets of network parameters.
Note, that for a ResNet, the number of trainable weight matrices equals the number of layers, whereas the spline-based network trains for $L+d$ weight matrices while the number of layers (time steps) is fixed.

\begin{figure}
    \centering
    \includegraphics[width=.49\textwidth]{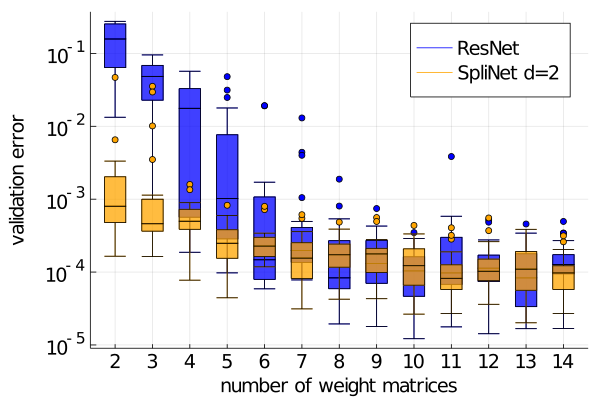}
    \includegraphics[width=.49\textwidth]{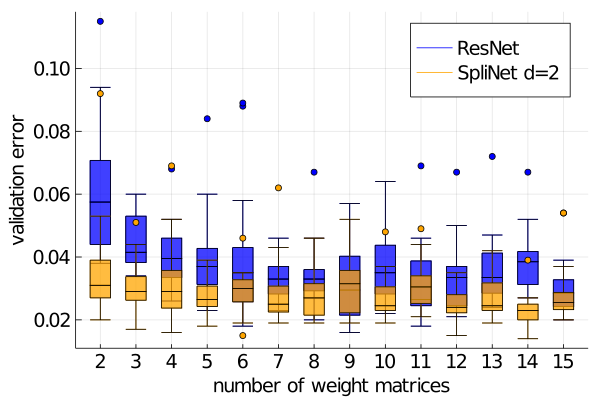}
    \caption{Median validation accuracy with lower and upper quartiles over numbers $(L+d)$ of trainable network parameter weights and biases for the $\sin(5x)$ (left) and the Peaks test case (right).}
    \label{fig:varyingL_boxplots}
\end{figure}

\begin{figure}
    \centering
    \includegraphics[width=.49\textwidth]{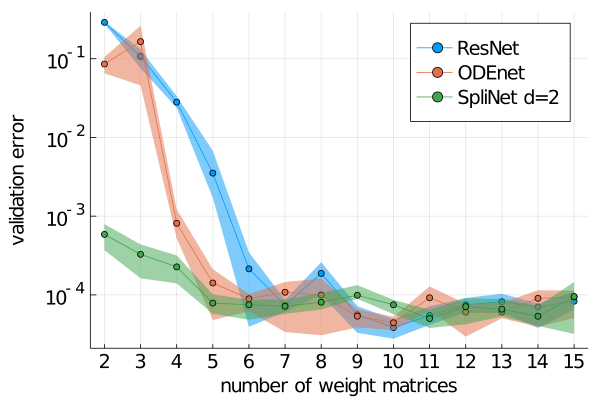}
    \includegraphics[width=.49\textwidth]{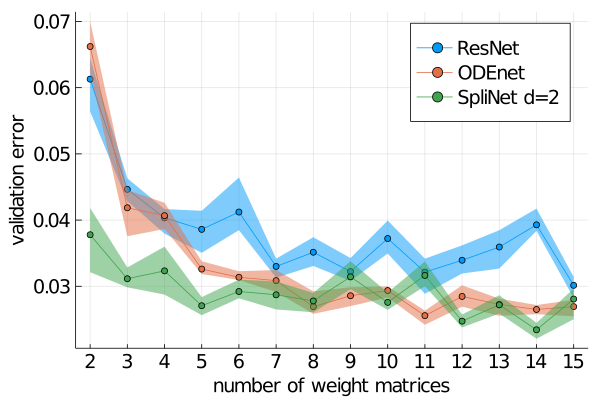}
    \caption{Mean validation accuracy with $95\%$ confidence intervals over numbers $(L+d)$ of trainable network parameter sets $\omega_l, \beta_l$, for the $\sin(5x)$ (left) and the Peaks test case (right).}
    \label{fig:varyingL_mean}
\end{figure}

For various $L$, Figure \ref{fig:varyingL_boxplots} plots the median validation error with lower and upper $25\%$ quartiles for the $\sin(5x)$ (left) and the Peaks test case (right), gathered over 260 training runs. We observe that the spline-based network yields greater accuracy most notably in the low-layer regime indicating the reduction in trainable network parameters. In particular, both the median as well as the best validation error for the SpliNet are drastically reduced for SpliNet on the $\sin(5x)$ test case, such as for $L=2$ layers in a ResNet, the same number of trainable parameters in a SpliNet reduce the median and minimum error by two orders of magnitude. Similarly, the Peaks test case shows lower median and minimum validation error for any small $L$ whereas the ResNet requires 6 trainable layers to reach the same accuracy of $98\%$.
Similar observations can be made from Figure \ref{fig:varyingL_mean} showing mean validation accuracy together with $95\%$ confidence intervals for ResNet, ODEnet and a SpliNet of degree $d=2$.

\subsection{Impact of the time scale}
\label{sec:impact-time-scale}

To illustrate the importance of correctly inferring the time scale associated with the given problem, we consider a network with the $\tanh$ activation function.
We remark that for this example, it is important to choose an activation function that is not homogeneous (e.g.\ ReLU), since as noted above, in the homogeneous case the effect of the time scale is completely controlled by scaling the weights and biases.
The corresponding system of ordinary differential equations takes the form
\[
  \frac{\partial x(t)}{\partial t} = \tanh(W(t)x(t) + b(t)), \qquad t \in (0,T),
\]
where $W(t)$ is a linear transformation, and $b(t)$ is a bias vector, both parameterized by B-spline basis functions whose coefficients it is our goal to learn.
As above, we rescale the pseudo-time variable to pose the problem over the reference domain, $\tref \in (0,1)$,
\[
  \frac{\partial \xref(\tref)}{\partial \tref} = \lambda \tanh(\hat{W}(\tref)\xref(\tref) + \hat{b}(\tref)), \qquad \tref \in (0,1).
\]
Since we have $-1 \leq \tanh(\varphi) \leq 1$ for all $\varphi$, we can infer the bounds for the solution $\xref$ at $t=1$,
\[
  \xref(0) - \lambda \leq \xref(1) \leq \xref(0) + \lambda.
\]
As a consequence, the choice of $\lambda$ determines bounds on the magnitude of the solution.

\begin{figure}
  \centering
  \includegraphics[width=0.6\linewidth]{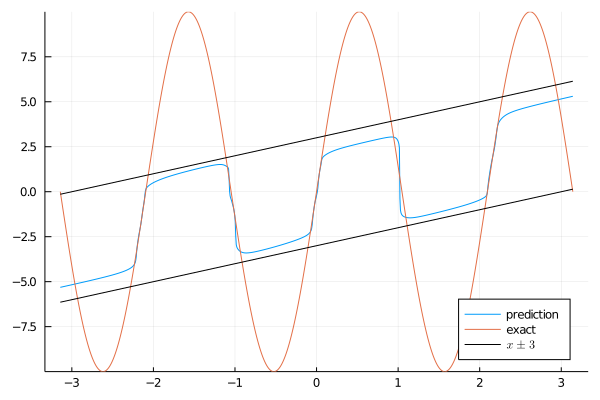}
  \caption{
    Learning $10\sin(3x)$ with fixed time scale parameter of $\lambda=3$ using $\tanh$ activation function.
    Predicted values are bound by $x \pm \lambda$, limiting the accuracy in cases where the time scale is not well-suited for the given problem.
    }
  \label{fig:sine-bounds}
\end{figure}

In this example, we wish to learn the scaled sine function $10 \sin(x)$ for $x \in [-\pi,\pi]$.
For a first test case, we fix $\lambda = 3$, and repeat the learning process described above.
In this case, the training and validation accuracy reaches a maximum of about 80\%; this is in contrast to the previous test cases, for which accuracy of above 98\% is easily achievable.
In Figure \ref{fig:sine-bounds} we visualize the predicted solution together with the bounds $x \pm \lambda$.
From this figure, it is clear that the bounds on the magnitude of the solution imposed by the fixed time scale $\lambda=3$ prevent the solution from accurately approximating the target function.

To remedy this issue, instead of choosing the time scale $\lambda$ to be a fixed parameter, we instead \textit{learn} the time scale by choosing the control function $\theta(\tref) = (\lambda, \hat{W}(\tref), \hat{b}(\tref))$, so that an appropriate value of $\lambda$ is chosen during the optimization process.
We perform 100 training runs, each for 200 epochs, with random initialization of the weights and biases, and set the initial $\lambda = 3$.
Since $\lambda$ is now a learnable parameter, the network is able to accurately approximate the target function.
Of the 100 runs, the average learned time scale was $\lambda = 13.59$, with values ranging from 10.72 to 18.47.
The average accuracy for these cases was 98.8\%.

\section{Conclusion}

This paper introduces a class of networks that uses B-spline basis functions to parameterize the network weights and biases (controls) across layers of a neural ODE network.  Instead of equipping each layer of a discretized ODE network with a set of trainable parameters, the spline-based network (SpliNet) trains for the coefficients of a certain number of B-spline basis functions.
The spline-based ODE network can be evaluated at any time point for any discretization of the underlying network ODE.
An ODE discretization choose the discretization based on a desired
accuracy, most notably tuned by increasing the number of time steps / layers and decreasing the time step size to ensure stable forward propagation.
While for a standard ODE network this implies an increase in the number of trainable parameters hence higher optimization complexity, the spline-based network achieves this for a fixed set of trainable parameters.

Our numerical results further demonstrate the benefits of B-spline based neural networks. Most importantly, we observe greater robustness with respect to hyperparameters such as network architecture, parameters for the training algorithm, and network initialization.
In comparison to a standard ODE network that discretizes each layer pointwise with a set of trainable weights and biases, as well as a standard residual neural network, the spline-based network performs better on average with higher mean and median accuracy and a tighter standard deviation on our test cases.

In the tests performed, the performance of the spline networks was not overly sensitive to the degree of the B-spline basis functions.
While higher degree basis functions increase smoothness of the network weights across the layers, we numerically observe that degree-one basis function  captures the benefits of spline-based network controls to increase regularity of the training problem.
Degree-one B-spline basis functions are ``hat functions'' that yield network weights that are piecewise linear across multiple layers. Hence, choosing as many basis functions as network layers recovers a standard ODEnet.
However, the number of knots for the hat-functions can be chosen independent of the network architecture and the discretization of the network ODE.
It can be adapted based on a desired training accuracy, rather than the number of layers that are needed to discretize the underlying network dynamics in order to guarantee stable forward and backward propagation of the underlying ODE.

Our numerical tests demonstrate that the spline-based parameterization of network weights allows for a reduction of the networks parameters while maintaining its approximation power when compared to standard ResNet training.
In fact, we observe almost constant validation accuracy even for a very small number of trainable parameters for degree-one spline-based network controls which can not be observed for ResNet or ODEnet training.

One additional benefit of using B-spline basis functions is that they are hierarchical and well suited for integration into multilevel or multigrid training such as \cite{gaedke2020multilevel, gunther2020layer}. Furthermore, the spline-based network controls allow for adaptive time stepping and different (higher-order) ODE network discretizations, which will be the topic of future work.

\section*{Acknowledgments}

This work was performed under the auspices of the U.S.\ Department of Energy by Lawrence Livermore National Laboratory under Contract DE-AC52-07NA27344 (LLNL-JRNL-819654).
This document was prepared as an account of work sponsored by an agency of the United States government.
Neither the United States government nor Lawrence Livermore National Security, LLC, nor any of their employees makes any warranty, expressed or implied, or assumes any legal liability or responsibility for the accuracy, completeness, or usefulness of any information, apparatus, product, or process disclosed, or represents that its use would not infringe privately owned rights.
Reference herein to any specific commercial product, process, or service by trade name, trademark, manufacturer, or otherwise does not necessarily constitute or imply its endorsement, recommendation, or favoring by the United States government or Lawrence Livermore National Security, LLC.
The views and opinions of authors expressed herein do not necessarily state or reflect those of the United States government or Lawrence Livermore National Security, LLC, and shall not be used for advertising or product endorsement purposes.

\bibliographystyle{siamplain}
\bibliography{references}

\end{document}